\documentclass[sigconf]{acmart}

\AtBeginDocument{%
  \providecommand\BibTeX{{%
    \normalfont B\kern-0.5em{\scshape i\kern-0.25em b}\kern-0.8em\TeX}}}

\copyrightyear{2022} 
\acmYear{2022} 
\setcopyright{acmlicensed}\acmConference[KDD '22]{Proceedings of the 28th ACM SIGKDD Conference on Knowledge Discovery and Data Mining}{August 14--18, 2022}{Washington, DC, USA}
\acmBooktitle{Proceedings of the 28th ACM SIGKDD Conference on Knowledge Discovery and Data Mining (KDD '22), August 14--18, 2022, Washington, DC, USA}
\acmPrice{15.00}
\acmDOI{10.1145/3534678.3539372}
\acmISBN{978-1-4503-9385-0/22/08}
\usepackage{subcaption}
\usepackage{booktabs}
\usepackage{bm}
\usepackage{bbm}
\usepackage{tcolorbox}
\usepackage{multicol,multirow}
\usepackage{natbib}
\usepackage{comment}
\usepackage{booktabs}

\usepackage{enumerate}   
\usepackage{amsmath}
\usepackage{tikz}

\usetikzlibrary{positioning,arrows}

\usepackage{wrapfig}

\DeclareMathOperator*{\argmin}{arg\,min}

\usepackage{amsmath}
\usepackage{appendix}

\usepackage{amsthm}
\usepackage[capitalise]{cleveref}
\Crefname{assumption}{Assumption}{Assumptions}

\theoremstyle{plain}
\newtheorem{theorem}{Theorem}
\newtheorem{lemma}{Lemma}

\theoremstyle{definition}
\newtheorem{assumption}{Assumption}
\newtheorem{proposition}{Proposition}
\newtheorem{definition}{Definition}
\newtheorem{remark}{Remark}

\newcommand{\iid}{ \stackrel{\mathrm{i.i.d.}}{\sim} }

\newcommand{\sign}{\mathrm{sign}}

\newcommand{\E}{\mathbb{E}}

\newcommand{\pa}{\mathrm{\pa}}

\renewcommand{\eqref}[1]{(\ref{#1})}

\newcommand{\RN}[1]{%
  \textup{\uppercase\expandafter{\romannumeral#1}}%
}

\usepackage{color,lscape,longtable}

\def\boxit#1{\vbox{\hrule\hbox{\vrule\kern6pt\vbox{\kern6pt#1\kern6pt}\kern6pt\vrule}\hrule}}

\usepackage{xcolor}
\newcount\Comments  
\newcommand{\kibitz}[2]{\ifnum\Comments=1\textcolor{#1}{#2}\fi}

\newcommand{\annot}[2]{\underbrace{#1}_{\text{#2}}}
\usepackage{todonotes}

\def \Re {\mathbb{R}}
\def \Na {\mathbb{N}}

\def \E {\mathbb{E}}
\def \hE {\hat{\E}}
\def \F {\mathcal{F}}

\def \f {f}

\def \supf {\sup_{\f \in \F}}

\def \loss {\ell}

\newcommand{\rClass}{\mathcal{H}}

\newcommand{\Rademacher}[2]{\Rad_{#1}^{#2}}

\def \rad {\sigma}
\def \ERad {\E_{\rad}}
\def \Rad {\mathcal{R}}

\def \EX {\E}

\def \hEjk {\hE_{(i, j)}}

\def \X {X}

\def \hS {\hat{S}}
\def \EhS {S}
\def \ljk {\loss_{(j, k)}}
\def \sumj {\sum_{j=1}^J}
\def \sumjk {\sum_{k=1}^{K_j}}
\def \Liprhoj {L_{\rho_j}}
\def \LipLossjk {L_{\ljk}}

\def \njk {n_{(j, k)}}
\def \pjk {p_{(j, k)}}
\def \InSpace {\mathcal{X}}

\def \Xsetjk {\mathcal{D}_{(j, k)}}

\usepackage{bbm} 

\newcommand{\Radnp}{\Rademacher{n}{p}}

\usepackage{braket}
\usepackage{algorithm}
\usepackage{algorithmic}
\usepackage{newfloat}
\usepackage{listings}
\begin{document}
\fancyhead{}

\title{Learning Classifiers under Delayed Feedback with a Time Window Assumption}


\allowdisplaybreaks

\author{Shota Yasui}
\authornote{Both authors contributed equally to this research.}
\email{yasui_shota@cyberagent.co.jp}
\orcid{0000-0002-0076-5965}
\affiliation{%
  \institution{CyberAgent, Inc.}
  \city{Tokyo}
  \country{Japan}
}

\author{Masahiro Kato}
\authornotemark[1]
\email{masahiro_kato@cyberagent.co.jp}
\orcid{0000-0001-7090-5735}
\affiliation{%
  \institution{CyberAgent, Inc.}
  \city{Tokyo}
  \country{Japan}
}

\begin{abstract}
We consider training a binary classifier under delayed feedback (\emph{DF learning}). For example, in the conversion prediction in online ads, we initially receive negative samples that clicked the ads but did not buy an item; subsequently, some samples among them buy an item then change to positive. In the setting of DF learning, we observe samples over time, then learn a classifier at some point. We initially receive negative samples; subsequently, some samples among them change to positive. This problem is conceivable in various real-world applications such as online advertisements, where the user action takes place long after the first click. Owing to the delayed feedback, naive classification of the positive and negative samples returns a biased classifier. One solution is to use samples that have been observed for more than a certain time window assuming these samples are correctly labeled. However, existing studies reported that simply using a subset of all samples based on the time window assumption does not perform well, and that using all samples along with the time window assumption improves empirical performance. We extend these existing studies and propose a method with the unbiased and convex empirical risk that is constructed from all samples under the time window assumption. To demonstrate the soundness of the proposed method, we provide experimental results on a synthetic and open dataset that is the real traffic log datasets in online advertising.
\end{abstract}



\maketitle

\section{Introduction}
Let us consider the problem of training a classifier under delayed feedback (\emph{DF Learning}). For example, we train a classifier for conversion prediction in online advertisements. For conversion prediction, we need to predict whether a certain ad click will lead to the purchase of some items on the advertiser's service. If a purchase occurs after the click, it is treated as a positive sample; if no purchase occurs, it is treated as a negative sample. There is a certain time window between the click and the purchase since users need time to make a decision to purchase. Such a time window can cause mislabeling in the training data. This is because a user who clicked on an ad just before the end of data collection may not have decided to purchase the product until the end of data collection. As a result, the part of training data, especially recently observed, is mislabeled, and then the classifier trained in the dataset will be suffered from the bias and deteriorate its performance.
The problem of DF learning arises in various cases such as medical treatment assignment, recommendation, and distributed learning \citep{Agarwal2011delay,Chapelle2014,Zhou2017delay,Yoshikawa2018,Burke2018,yasui2020,badanidiyuru2021handling, saito_dual, Yang_Li_Han_Zhuang_Zhan_Zeng_Tong_2021, asympt_unbiased, follow_prophet, reward_modification}. 

We can classify the methods of DF learning into online and offline prediction settings. Online DF learning includes sequential parameter updating \citep{Ktena2019,defer}, online learning \cite{pmlr-v28-joulani13}, and multi-armed bandit problem \citep{Quanrud2015,Bianchi2019delay,Zhou2019learning,Bistritz2019, delayed_bandit_han}. We focus on offline DF learning and do not update the parameter online. Solutions involving offline DF learning follow two main approaches. The first is to assume that a sufficiently long time window reduces the bias between the observed label and its ground truth \citep{Xinran2014}. This assumption is due to the fact that the labels of most samples that have been observed for a certain amount of time are observed correctly. Based on this assumption, \citet{Xinran2014} proposed a naive logistic regression, and \citet{yasui2020} proposed an importance weighting (IW) method for constructing a consistent empirical risk. The second approach is to specify a probabilistic model of the delayed feedback \citep{Chapelle2014, Yoshikawa2018}. Due to the poor empirical performance of the latter approach and difficulty in model specification, this study adopts the first approach.

In this paper, we propose a new approach for offline DF learning. The proposed method approximates the same population risk with \citet{yasui2020} in an end-to-end manner. In addition, our method is based on convex optimization and provides theoretical guarantees regarding the estimation error. Furthermore, to improve the empirical performance, we further provide a non-negative correction to the empirical risk, following the approaches by \citet{kiryo2017} and \citet{kato2020dre}.

Four main contributions are made: (i) proposing a novel convex empirical minimization for DF learning with a time window and stationarity assumption; (ii)  providing a non-negative correction to the original convex empirical risk minimization (ERM) for using a flexible model; (iii) demonstrating the effectiveness of the proposed method using both synthetic data and real-world log data \citep{Chapelle2014}; (iv) the existing studies are summarized in a unified manner from the assumptions perspective.

\section{Problem Setting}
We consider the problem setting introduced by \citet{Chapelle2014}. For a sample $i\in\mathbb{N}$ with a feature $X_i\in\mathcal{X}$, we consider a binary classification problem to classify $X_i$ into one of the two classes $\{-1, +1\}$. Let a classifier $g:\mathcal{X}\to \mathbb{R}$ be a function that assigns a label $\hat{C}_i$ to an individual with a feature $X_i$ such that $\hat{C}_i = \sign (g(X_i))$. We assume that there exists a joint distribution $p(X_i, C_i)$, where $C_i \in \{-1, +1\}$ is the class label of $X_i$. In DF learning, depending on applications, two goals are considered. The first goal is to train the Bayes optimal classifier, which minimizes the population classification risk $J_{\mathrm{0}\mathchar`-\mathrm{1}}(g)$ defined as 
\[J_{\mathrm{0}\mathchar`-\mathrm{1}}(g) = \gamma\mathbb{E}_{C=+1}[\ell_{\mathrm{0}\mathchar`-\mathrm{1}}(g(X_i))] + (1-\gamma)\mathbb{E}_{C=-1}[\ell_{\mathrm{0}\mathchar`-\mathrm{1}}(-g(X_i))],\] where $\mathbb{E}_{C=c}$ denotes the expectation over $p(X_i\mid C_i=c)$, $\gamma  = p(C_i=+1)$, and $\ell_{\mathrm{0}\mathchar`-\mathrm{1}}$ is the zero-one loss $\ell_{\mathrm{0}\mathchar`-\mathrm{1}}(z)=\frac{1}{2}\sign(z) + \frac{1}{2}$. In practice, we replace $\ell_{\mathrm{0}\mathchar`-\mathrm{1}}$ with a surrogate loss, such as the logistic loss. The population is denoted with a surrogate loss function $\ell$ as $J(g)$. For ease of discussion, the formulation with the surrogate loss $\ell$ is mainly considered in the following sections. For a set of measurable functions $\mathcal{F}$, the optimal classifier $g^*$ is defined as 
\[g^*_{} =\argmin_{g\in\mathcal{F}}J_{\mathrm{0}\mathchar`-\mathrm{1}}(g).\] The second goal is to estimate the conditional probability $p(C_i\mid X_i)$. When using the zero-one loss or other specific losses such as logistic loss, the minimizer $g^*$ coincides with $p(C_i\mid X_i)$. In various applications, we have more interest on an estimate of $p(C_i\mid X_i)$ rather than the prediction results. For example, in online advertisement, by using $p(C_i\mid X_i)$, we decide the bid price as Eq.~(1) of \citet{Chapelle2014}. Let us note that the first and second goals are closely related. 

\begin{remark}
\label{rem:stationarity_point}
For specific loss functions, $g^*(X_i)$ is equal to $p(C_i\mid X_i)$. For example, when using the logistic loss, we obtain $p(C_i\mid X_i)$ as the minimizer of the population risk.
\end{remark}

\subsection{Data Generating Process}
In our setting, in the time series $[T]=\{1,2,\dots,T\}$, we obtain a sample $i\in\mathbb{N}$ with the feature $X_i$ at an arriving time $A_i$ sequentially, where $T$ is the period we train a classifier. For the sample $i$, instead of observing the true class $C_i$ directly, a temporal label $Y_i(e) \in \{-1, +1\}$ reveals at each elapsed time $e \in \{1,\dots,T-A_i\}$ after arriving time $A_i$. In general, $Y_i(e)$ is reproducible if we preserve the timestamp when the label was observed in addition to $A_i$. Once we observe $Y_i(e)=+1$, the label $Y_i(s)$ is permanently $+1$ for all $s\geq e$, i.e., $Y_i(s) = C_i$ for all $s\geq e$. For example, let us assume a user clicked the ad at time $A_i$ and the user purchase the item 60 minutes after the click. This means that when the elapsed time $e$ is less than 60 minutes, the temporal label is still negative, i.e., $Y_i(e) = -1$. On the other hand, we observe $Y_i(e)=+1$, when $e$ is larger than 60 minutes.

Then, we describe a more formal data-generating process. For each sample $i\in\mathbb{N}$, at the $T$-th period, we obtain a dataset 
\[\left\{(X_i, \{Y_i(e)\}^{T-A_i}_{e=1}, A_i)\right\}^N_{i=1},\]
where $Y_i(e)$ is a temporal class label of $i$ at elapsed time $e$, and $A_i$ is the arrival time. We denote the period from $A_i$ to $t$ as $E^t_{A_i}$, which is the elapsed time after observing a sample $i$ at $A_i$, i.e., $E^t_{A_i} = t - A_i$ for $A_i \leq t \leq T$. To simplify the notation, $E^t_{A_i}$ is written as $E^t_{i}$ in the rest of the paper. We assume that the triple $\left(X_i, E^t_{i}, Y_i(E^t_{i})\right)$ is generated as follows:
\begin{align*}
(X_i, E^t_{i}, Y_i(E^t_{i})) \iid & p\left(X_i, E^t_{i}, Y_i(E^t_{i})\right) \\\nonumber
&= p\big(X_i, E^t_{i}\big)p\big(Y_i(E^t_{i})\mid X_i, E^t_{i}\big) \\\nonumber
&= p\big(X_i\big)p\big(E^t_{i}\big)p\big(Y_i(E^t_{i})\mid X_i, E^t_{i}\big),
\end{align*}
where $p(X_i, E^t_{i})$, $p\big(E^t_{i}\big)$, and $p(X_i)$ are the probability densities of $(X_i, E^t_{i})$, $E^t_{i}$, and $X_i$, respectively, and $p\big(Y_i(E^t_{i})\mid X_i, E^t_{i}\big)$ is the conditional probability density of $Y_i(E^t_{i})$ given $X_i$ and $E^t_{i}$. Here, we assume that $p\big(X_i| E^t_{i}\big) = p(X_i)$ for all $t>A_i$, that is, a sample feature is not dependent on the period. We denote the dataset $\left\{\left(X_i, Y_i(E^t_{i}), E^t_{i}\right)\right\}^N_{i=1}$ by $\mathcal{D}$.

\subsection{Time Window and Stationarity Assumptions and Oracle Datasets}
As well as \citet{yasui2020}, we introduce a deadline $\tau\in[T]$. For this deadline, we assume that a sample after spending $\tau$ period from the first observation $A_i$ has the correct label $C_i$, i.e., $Y_i(E^t_{i}) = C_i$ for $E^t_{i} \geq \tau$. Let us also define a label $S_i(E^t_{i}) \in \{-1, +1\}$, which indicates whether a temporal label $Y_i(E^t_{i})$ observed at $t$-th period is equal to $C_i$, i.e., $S_i(E^t_{i}) = +1$ if $Y_i(E^t_{i}) = C_i$; $S_i(E^t_{i}) = -1$ if $Y_i(E^t_{i}) \neq C_i$.

\begin{assumption}[Time Window Assumption]
\label{assumption:time_window}
$Y_i(E^t_{i}) = C_i$ for $E^t_{i} \geq \tau \Leftrightarrow A_i \leq t - \tau$.
\end{assumption}

We also assume that the conditional probability of the temporal labels is the same between different periods.

\begin{assumption}[Stationarity Assumption]
For all $i, j\in[N]$, $s\in[T]$, $X\in \mathcal{X}$, and $t'\in[T]$,
\begin{align*}
p\big(Y_i\big(E^t_{i}\big)\mid X_{i} = X, E^t_{i}=s\big)=p\big(Y_j\big(E^{t'}_{j}\big)\mid X_{j} = X, E^{t'}_{j}=s\big).
\end{align*}
\end{assumption}

Under the time window assumption, we reconstruct oracle datasets from the original dataset $\mathcal{D}$. Assume that $\tau \leq \lfloor T/2 \rfloor$. Under the time window assumption, we construct the oracle dataset $\mathcal{E}=\left\{\left(X_{j}, C_{j}, S_{j}(E^{T-\tau}_{j})\right)\right\}^M_{j=1}$ from samples that are observed over $\tau$ periods, i.e., 
\[\tau < E^t_{j} \leq T \Leftrightarrow 0\leq A_j \leq T - \tau,\] where $S_j(E^{T-\tau}_{j}) \in \{-1, +1\}$ is assigned $+1$ if $Y_j(E^{T-\tau}_{j}) = Y_j(E^t_{j}) = C_j$.

\section{Unbiased Formulation of DF learning using the Time Window Assumption}
\label{sec:unb_dfl}
In this section, we first organize the relationships among random variables in DF learning. Then we define the unbiased risk in DF learning under the time window and the stationary assumption. At last, we introduce our proposed methods.

\subsection{Relationship among Random Variables}
\label{sec:relation_randomvar}
To construct a risk estimator, we investigate the relationship among random variables $X_i$, $Y_i(E^t_{i})$, $C_i$, $E^t_{i}$, and $S_i(E^t_{i})$. \citet{yasui2020} found the following relationship. The samples labeled as $Y_i(E^t_{i})=+1$ in the biased dataset $\mathcal{D}$ are true positive ($C_i=+1$). Therefore, $Y_i(E^t_{i}) = +1 \Leftrightarrow S_i(E^t_{i}) =+1, C_i =+1$. Under delayed feedback, however, some positive samples ($C_i =+1$) are mislabeled $(S_i(E^t_{i}) =-1)$. Hence, the negative samples in biased dataset $\mathcal{D}$ consist of false and true ones. Formally, 
\begin{align*}
Y_i(E^t_{i})=-1\Leftrightarrow C_i=-1\ \ \ \mathrm{or}\ \ \  S_i(E^t_{i})=-1.
\end{align*}
Based on these observations, the relationships between the conditional distributions of $Y_i(E^t_{i})$ and $C_i$ are given as:
\begin{align*}
p(Y_i(E^t_{i})=+1\mid Z^T_i) =&  p(C_i=+1, S_i(E^t_{i})=+1\mid Z^T_i),\\
p(Y_i(E^t_{i})=-1\mid Z^T_i)=& p(C_i=-1\mid Z^T_i) \\
&+ p(C_i=+1, S_i(E^t_{i})=-1\mid Z^T_i),
\end{align*}
where we denote $(X_i, E^t_{i})$ as $Z^T_i$. Since the true positive samples contain both correctly and incorrectly observed samples, we can obtain
\begin{align}
\label{eq:CYrel}
    &p(Z^T_i, C_i=+1) \nonumber\\
    &= p(Z^T_i, C_i=+1, S_i(E^t_{i})=+1)\nonumber\\
    &\ \ \ + p(Z^T_i, C_i=+1, S_i(E^t_{i})=-1) \nonumber\\
    \Leftrightarrow &p(C_i=+1)p(Z^T_i\mid C_i=+1) \nonumber\\
    &= \pi p(Z^T_i\mid C_i=+1, S_i(E^t_{i})=+1)\nonumber\\
    &\ \ \ + \zeta p(Z^T_i\mid C_i=+1, S_i(E^t_{i})=-1),
\end{align}
where 
\begin{align*}
    \pi = p(C_i=+1, S_i(E^t_{i})=+1) = p(Y_i(E^t_{i})=+1)
\end{align*}
and
\begin{align*}
    \zeta = p(C_i=+1, S_i(E^t_{i})=-1).
\end{align*}
By using Eq~\eqref{eq:CYrel}, we can obtain the gap between the density in a positive biased dataset and the positive ideal dataset as follows:
\begin{align}
\label{eq:prob1}
\gamma p(Z^T_i\mid C_i=+1) =& \pi p(Z^T_i\mid Y_i(E^t_{i})=+1) \nonumber\\
&\ \ \ + \zeta p(Z^T_i\mid C_i=+1, S_i(E^t_{i})=-1).
\end{align}
Similarly we can obtain the gap in the negative data as follows:
\begin{align}
\label{eq:prob2}
(1-\gamma)p(Z^T_i\mid C_i=-1)=&(1-\pi)p(Z^T_i\mid Y_i(E^t_{i})=-1) \nonumber\\
&\ \ \ - \zeta p(Z^T_i\mid C_i=+1, S_i(E^t_{i})=-1).
\end{align}

\subsection{Construction of Unbiased Risk Estimator}
\label{sec:construction_unbiased_risk}
Let us consider directly using $\mathcal{D}$ for the binary classification loss. The population risk of $\mathcal{D}$ is defined as 
\[J^{\mathrm{BL}}(g) = \mathbb{E}\left[\ell\big(Y_i(E^t_{i})g(X)\big)\right],\] and we denote its empirical version as $\widehat{J}^{\mathrm{BL}}(g)$, where BL represents Biased Logistic regression. Note that because the true label $C_i$ is independent of $E^t_{i}$, we can construct a classifier using only $X_i$. Since the BL uses biased data set $\mathcal{D}$, the risk used in BL $J^{\mathrm{BL}}(g)$ is not equivalent to $J(g)$ and biased. 

We can correct the bias of $J^{\mathrm{BL}}(g)$ by using the relationships shown in Eq.~\eqref{eq:prob1} and Eq.~\eqref{eq:prob2} as follows:
\begin{align*}
J(g) =& \gamma\mathbb{E}\left[\ell\big(g(X_i)\big)\right] + (1-\gamma) \mathbb{E}\left[\ell\big(-g(X_i)\big)\right]\\\nonumber 
=&J^{\mathrm{BL}}(g) + \zeta \mathbb{E}_{S=-1, C=+1}[\ell(g(X_i))]\\\nonumber
&- \zeta \mathbb{E}_{S=-1, C=+1}[\ell(-g(X_i))],
\end{align*}
where $\mathbb{E}_{S=-1, C=+1}$ denotes the expectation over $p(Z^T_i \mid S_i(E^t_{i})=-1, C_i=+1)$. Intuitively, adding $\zeta \mathbb{E}_{S=-1, C=+1}[\ell(g(X_i))]$ to $J^{\mathrm{BL}}(g)$ corrects for the positive loss in data where $C = 1$ and $S = -1$, and similarly subtracting $\zeta \mathbb{E}_{S=-1, C=+1}[\ell(-g(X_i))]$ corrects for the negative loss. Here, we used $\mathbb{E}_{Z|W}[g(X_i)] =\mathbb{E}_{X|W}[g(X_i)]$, where $\mathbb{E}_{Z|W}$ and $\mathbb{E}_{X|W}$ denote the expectations over  $p(Z^T_i \mid W_i)$ and $p(X_i \mid W_i=w)$ for a random variable $W_i$, respectively. Under this equivalent transformation, we can then obtain the empirical risk estimator using both $\mathcal{D}$ and $\mathcal{E}$:
\begin{align}
\label{eq:empirical_risk}
\widehat{J}(g) =& \frac{1}{N}\sum_{i\in\mathcal{D}} \ell\Big(Y_i\left(E^t_{i}\right)g(X_i)\Big)\\\nonumber 
&+ \frac{1}{M}\sum_{j\in\mathcal{E}}\mathbbm{1}\left[\Big(S_j(E^{T-\tau}_{j} )=-1\Big) \land \Big(C_j=+1\Big)\right]\widetilde{\ell}\Big(g(X_j)\Big),\nonumber
\end{align}
where $\tilde{\ell}(g(x)) = \ell\big(g(x)\big) - \ell\big(-g(x)\big)$, and we call it a composite loss. By using $T-\tau \geq \tau$ and the stationarity assumption, the second term on the right hand side(RHS) converges to $\mathbb{E}_{S=-1, C=+1}[\ell(-g(X_i))]$, where we used $p\left(X_j, S_j(u)=-1, C_j=+1\right) = 0$ for $u\geq \tau$ from the time window assumption. This empirical risk is unbiased for $J(g)$. Note that to approximate the expectation, the support of $E^T_j-\tau$ should be larger than that of $E^t_{i}$ for $\mathcal{D}$.
\subsection{Proposed Estimators}
Here, we introduce our proposed methods convDF and nnDF. Firstly, we proposed convex DF learning (convDF) that minimize the unbiased loss $\widehat{J}(g)$ proposed in section \ref{sec:construction_unbiased_risk} by the ERM. However, when the hypothesis class is large, the ERM of $\widehat{J}(g)$ causes overfitting, as reported by \citep{kiryo2017} due to the form of the empirical risk. Therefore, we propose to use a non-negative modification of the loss. Denote the positive and negative parts of the empirical risk as $\widehat{J}^{(+)}(g)$ and $\widehat{J}^{(-)}(g)$, respectively. Then,  Eq.~\eqref{eq:empirical_risk} yield the following relationship:
\begin{align*}
    \widehat{J}^{(+)}(g) =& \underbrace{\frac{1}{N}\sum_{i\in\mathcal{D}}\mathbbm{1}\left[Y_i\left(E^t_{i}\right)=+1\right]\ell\Big(g(X_i)\Big)}_{\widehat{J}^{(+)}_{\mathcal{D}}(g)} \\\
    &+ \underbrace{\frac{1}{M}\sum_{j\in\mathcal{E}}\mathbbm{1}\left[\Big(S_j(E^{T-\tau}_{j} )=-1\Big) \land \Big(C_j=+1\Big)\right]\ell\Big(g(X_j)\Big)}_{\widehat{J}^{(+)}_{\mathcal{E}}(g)}\\\
    \widehat{J}^{(-)}(g) =&
    \underbrace{\frac{1}{N}\sum_{i\in\mathcal{D}}\mathbbm{1}\left[Y_i\left(E^t_{i}\right)=-1\right]\ell\Big(-g(X_i)\Big)}_{\widehat{J}^{(-)}_{\mathcal{D}}(g)} \\\
    &- \underbrace{\frac{1}{M}\sum_{j\in\mathcal{E}}\mathbbm{1}\left[\Big(S_j(E^{T-\tau}_{j} )=-1\Big) \land \Big(C_j=+1\Big)\right]\ell\Big(-g(X_j)\Big)}_{\widehat{J}^{(-)}_{\mathcal{E}}(g)}
\end{align*}

In $\widehat{J}^{(-)}(g)$, the empirical minimization leads $-\widehat{J}^{(-)}_{\mathcal{E}}(g)$ to $-\infty$ to minimize the overall empirical risk using Positive-Unlabeled Learning (PU Learning) \citep{kiryo2017} and density ratio estimation \citep{kato2020dre}. Therefore, we similarly propose using an alternative empirical risk with non-negative correction to the negative risk part as 
\[\widetilde{J}(g) = \widehat{J}^{(+)}_{\mathcal{D}}(g) + \widehat{J}^{(+)}_{\mathcal{E}}(g) + \min\left\{\widehat{J}^{(-)}_{\mathcal{D}}(g) - \widehat{J}^{(-)}_{\mathcal{E}}(g), 0\right\},\] then minimize to learn a classifier. We call this non-negative correction approach as non-negative DF learning (nnDF). For a function class $\mathcal{H}$, the classifiers of convDF and nnDF are $\hat{g}=\argmin_{g\in\mathcal{H}} \widehat{J}(g)$ and $\tilde{g}=\argmin_{g\in\mathcal{H}} \widetilde{J}(g)$ accordingly.

\subsection{Proposed Algorithm}
\label{sec:algorithm_theory}
Herein, we introduce the algorithms for convDF and nnDF. Firstly, we explain the convexity of the empirical loss. Secondly, we introduce the learning algorithm. Since the surrogate loss function $\ell(g(x))$ is convex, the empirical loss becomes convex if the composite loss $\tilde{\ell}(g(x))$ is convex. For the composite loss $\tilde{\ell}(z)$, Theorem~1 of \citet{duplessis2015} states that if the composite loss $\tilde{\ell}(g(x))$ is convex and $g(x)$ is a linear model, then  $\tilde{\ell}(g(x))$ is linear, that is, $\tilde{\ell}(z)=-g(x)$. When model $g(x)$ is linear, the composite loss is convex and the entire empirical loss is convex. In Table~1 of \citet{duplessis2015}, they summarize the surrogate loss functions. For example, when using the logistic loss, the empirical risk can be written as follows: 

\begin{align*}
\widehat{J}_{\mathrm{logistic}}(g) =& \frac{1}{N}\sum_{i\in\mathcal{D}} \log\left(1+\exp\Big(-Y_i\left(E^t_{i}\right)g(X_i)\Big)\right) \\\nonumber 
&\ \ \ - \frac{1}{M}\sum_{j\in\mathcal{E}}\mathbbm{1}\left[\Big(S_j(E^{T-\tau}_{j} )=-1\Big) \land \Big(C_j=+1\Big)\right]g(X_j).
\end{align*} 

Based on this result, we show the gradient of $\widehat{J}_{\mathrm{logistic}}(g)$, which is useful when training the classifier using a gradient-based optimization method in the \ref{appdx:gradients}. We also show the gradient of $\widetilde{J}_{\mathrm{logistic}}(g)$ of nnDF in the same appendix. These gradients are used in the learning algorithm we will introduce next.

Secondly, we explain the learning algorithm. In ERM, we jointly minimize the empirical risk and the regularization term denoted by $\mathcal{R}(g)$. We then train a model using gradient descent with a learning rate $\xi$ and regularization parameter $\lambda$. We can choose $\lambda$ based on cross-validation. When conducting gradient descent, we heuristically introduce the gradient descent/ascent algorithm as in \citet{kiryo2017}. We show the pseudo-algorithms for convDF and nnDF with and without the gradient descent/ascent algorithm in Algorithm~\ref{alg}. Although the theoretical details of the gradient descent/ascent algorithm are not discussed, the technique is known to improve performance when using flexible models such as neural networks. Note that the proposed algorithms are agnostic to the optimization procedure.
\begin{algorithm}[tb]
   \caption{convDF and nnDF}
   \label{alg}

\begin{algorithmic}
   \STATE {\bfseries Input.} The biased dataset $\mathcal{D}$, oracle dataset  $\mathcal{E}$, learning rate $\xi$, and the regularization coefficient $\lambda$. 
   \STATE {\bfseries Output.} An estimator of $p(C_i\mid X_i)$.
   \WHILE{No stopping criterion has been met.}
   \IF{convDF}
   \STATE Set gradient $\nabla \left\{\widehat{J}(g) + \lambda \mathcal{R}(g)\right\}$.
   \ELSE
   \IF{$\widehat{J}^{(-)}_{\mathcal{D}}(g)  \geq 0$.}
   \STATE Set gradient $\nabla \left\{\widehat{J}(g) + \lambda \mathcal{R}(g)\right\}$
   \ELSE 
   \IF{Gradient ascent}
   \STATE Set gradient $\nabla \left\{ - \widehat{J}^{(-)}(g) + \lambda \mathcal{R}(g)\right\}$.
   \ELSE
   \STATE Set gradient $\nabla \left\{ \widehat{J}^{(+)}(g) + \lambda \mathcal{R}(g)\right\}$.
   \ENDIF
   \ENDIF
   \STATE Update $g$ with the gradient and the learning rate $\xi$. 
   \ENDIF
   \ENDWHILE
\end{algorithmic}
\end{algorithm}

\section{Theoretical Analysis}
In this section, we introduce the theoretical analysis of our proposed methods. At first, we explain the bias and consistency of nnDF. Then we introduce the error bounds.

\subsection{Bias and Consistency of nnDF}
Since convDF directly minimize the unbiased risk $\widehat{J}(g)$, convDF is unbiased. On the other hand, the empirical risk $\widetilde{J}(g)$ of nnDF is biased because for a fixed $g\in\mathcal{F}$, we can show that $\widetilde{J}(g)\geq \widehat{J}(g)$ for any $(\mathcal{D}, \mathcal{E})$. A remaining question is whether $\widetilde{J}(g)$ is consistent. Following \citet{kiryo2017}, we prove its consistency in here. First, partition all possible realizations $(\mathcal{D}, \mathcal{E})$ into $\mathcal{A}(g)=\{(\mathcal{D}, \mathcal{E})\mid \widehat{J}^{(-)}(g) \geq 0\}$ and $\mathcal{B}(g)=\{(\mathcal{D}, \mathcal{E})\mid \widehat{J}^{(-)}(g) < 0\}$. Assume that $C_g > 0$ and $C_\ell > 0$ such that $\sup_{g\in\mathcal{G}}\|g\|_\infty \leq C_g$ and $\sup_{|t|\leq C_g}\max_x\ell(t)\leq C_\ell$.

\begin{lemma}
\label{lmm:bias_consistency}
The following three conditions are equivalent: (A) the probability measure of $\mathcal{B}(g)$ is non-zero; (B) $\widetilde{J}(g)$ differs from $\widehat{J}(g)$ with a non-zero probability over repeated sampling of $(\mathcal{D}, \mathcal{E})$; (C) the bias of $\widetilde{J}(g)$ is positive. In addition, by assuming that there is $\alpha > 0$ such that $\widehat{J}^{(-)} (g) \geq \alpha$, the probability measure of $\mathcal{B}(g)$, which is $\mathrm{Pr}\left(\mathcal{B}(g)\right)$, can be bounded by
\begin{align}
\label{eq:bias_prob_bound}
\mathrm{Pr}\left(\mathcal{B}(g)\right) \leq \exp\left(-2(\alpha^2/C_\ell)^2/\left(3/N + 1/M\right)\right).
\end{align}
\end{lemma}
Based on Lemma~1, we can show the exponential decay of both the bias and consistency. For convenience, let $\chi_{N, M} = \sqrt{3/N} + \sqrt{1/M}$.

\begin{theorem}[Bias and Consistency]
\label{thm:bias_consistency}
Assume that  and denote by $\Delta_g$ the RHS of Eq.~\eqref{eq:bias_prob_bound}. As $N, M\to\infty$, the bias of $\widetilde{J}(g)$ decays exponentially: $0\leq \mathbb{E}\left[\widetilde{J}(g)\right] - J(g) \leq C_\ell \Delta_g$.
Moreover, for any $\delta>0$, let $C_\delta = C_\ell\sqrt{2\log \big(2/\delta\big)}$, then we have with probability at least $1-\delta$,
\begin{align}
\label{eq:dev_bound}
\left|\widetilde{J}(g) - J(g)\right|\leq C_\delta\cdot \chi_{N,M} + C_\ell\Delta_g,
\end{align}
and with probability at least $1-\delta-\Delta_g$,
\begin{align}
\label{eq:dev2}
\left|\widetilde{J}(g)-J(g)\right|\leq  C_\delta\cdot \chi_{N,M}. 
\end{align}
\end{theorem} 
Theorem~\ref{thm:bias_consistency} implies that for a fixed $g$, $\widetilde{J}(g)\xrightarrow{\mathrm{p}} J(g)$ in $\mathrm{O}_p(\sqrt{3/N} + \sqrt{1/M})$. Further note that $M\leq N$. Thus, the empirical risk has $\sqrt{M}$-consistency, as does the central limit theorem.  

\begin{table*}[t]
    \centering
     \caption{Comparison of Methods for DF learning.}
      \label{tab:comparison}
    \scalebox{1}[1]{
    \begin{tabular}{|c|ccccccc|} \hline
        Method & Use of $\mathcal{D}$ & Use of $\mathcal{E}$  & Time Window & Stationarity & Model Specification & Unbiasedness & Consistency \\
         \hline
        convDF &  Use  & Use  & Assume & Assume &  & $\bigcirc$ & $\bigcirc$ \\
        nnDF &  Use  & Use  & Assume & Assume &  &  & $\bigcirc$ \\
        BL  &  Use  &   &  &  &  &  &  \\
        TW  &    &  Use & Assume &  &  & $\bigcirc$ & $\bigcirc$ \\
        PUTW  &  Use  &  Use & Assume &  &  & $\bigcirc$ & $\bigcirc$ \\
        FSIW  &  Use  & Use  & Assume & Assume &  &  & $\bigcirc$ \\
        DFM  &  Use  &   &  &  & Specify & $\bigcirc$ & $\bigcirc$ \\
        \hline
    \end{tabular}}
   \end{table*}

\subsection{Estimation Error Bounds}
Assume that $C_g > 0$ and $C_\ell > 0$ such that $\sup_{g\in\mathcal{H}}\|g\|_\infty \leq C_g$ and $\sup_{|x|\leq C_g}\ell(x)\leq C_\ell$. For any function class $\mathcal{H}$, given sets of samples $\mathcal{D}$ and $\mathcal{E}$, we define the  empirical Rademacher complexities as $\mathcal{R}_{\mathcal{D}}(\mathcal{H}):= \frac{1}{N}\mathbb{E}_{\sigma}\left[\sup_{g\in\mathcal{H}}\sum^N_{i=1}g(X_i)\right]$ and $\mathcal{R}_{\mathcal{E}}(\mathcal{H}):= \frac{1}{M}\mathbb{E}_{\sigma}\left[\sup_{g\in\mathcal{H}}\sum^M_{j=1}g(X_j)\right]$.
Then, the estimation errors of convDF and nnDF are given from the following theorem.

\begin{theorem}[Estimation Error Bound of convDF]
\label{thm:est_error_convDF}
Assume that $\mathcal{H}$ is closed under negation, i.e., $g\in\mathcal{H}$ if and only if $-g\in\mathcal{H}$. Then, for any $\delta > 0$, with probability at least $1-\delta$, $\widehat{J}(\hat{g}) - J(g^*) \leq 8C_\ell \mathcal{R}_\mathcal{D}(\mathcal{H}) + 8C_\ell \mathcal{R}_\mathcal{E}(\mathcal{H})+2C_\delta\cdot \chi_{N,M}$.
\end{theorem}

\begin{theorem}[Estimation Error Bound of nnDF]
\label{thm:est_error_nnDF}
Assume that (a) $\inf_{g\in\mathcal{F}}(g)\geq \alpha > 0$ and denote by $\Delta$ the RHS of Eq.~\eqref{eq:bias_prob_bound}; (b) $\mathcal{H}$ is closed under negation, i.e., $g\in\mathcal{H}$ if and only if $-g\in\mathcal{H}$. Then, for any $\delta > 0$, with probability at least $1-\delta$, $\widetilde{J}(\tilde{g}) - J(g^*) \leq 16C_\ell \mathcal{R}_\mathcal{D}(\mathcal{H}) + 16C_\ell \mathcal{R}_\mathcal{E}(\mathcal{H})+2C_\delta\cdot \chi_{N,M} + 2C_\ell \Delta$.
\end{theorem}

\section{Related Work}
In this section, we review related work regarding DF learning in the problem setting we introduced. Each study defined DF learning independently, and relations are not clear. Therefore, we explain each study based on a unified manner. Firstly, we can naively apply a logistic regression to the biased dataset $\mathcal{D}$, wherein the empirical risk is written as $\widehat{J}^{\mathrm{BL}}(g) = \frac{1}{N}\sum_{i\in\mathcal{D}} \ell\Big(Y_i(E^t_{i})g(X_i)\Big)$. The minimizer of $\mathbb{E}\left[\widehat{J}^{\mathrm{BL}}(g)\right]$ is equal to $p(Y_i(E^t_{i})\mid X_i)$, which is biased from $p(C_i\mid X_i)$. We call this method \emph{biased logistic regression} (BL).

Secondly, we introduce methods using the time window assumption. To mitigate the bias, \citet{Xinran2014} proposed using a time window that is sufficiently long to reduce the bias between the label $Y_i(E^t_{i})$ and the ground truth $C_i$. They proposed \emph{Time Window regression} (TW) that minimizes an empirical risk defined as \[\widehat{J}^{\mathrm{TW}}(g) = \frac{1}{M}\sum_{j\in\mathcal{E}} \ell\big(C_jg(X_j)\big)\] in the oracle dataset $\mathcal{E}$. 

Under the time window assumption, it is also possible to formalize the problem of DF learning as PU learning. We can regard the positive data in the dataset $\mathcal{E}$ as the true positive data. Then, by considering all data in $\mathcal{D}$ as the unlabeled data, we can construct an empirical risk using convex PU learning as 
\begin{align*}
\widehat{J}^{\mathrm{PUTW}}(g) =& \frac{1}{M}\sum_{j\in\mathcal{E}} \mathbbm{1}\big[C_j = +1\big]\ell\big(-g(X_j)\big) \\
&-\frac{1}{M}\sum_{j\in\mathcal{E}} \mathbbm{1}\big[C_j = +1\big]\ell\big(-g(X_j)\big) + \frac{1}{N}\sum_{i\in\mathcal{D}} \ell\big(-g(X_i)\big).
\end{align*}
We call this PU learning approach as PUTW. In addition, because true negative data also exist in $\mathcal{E}$, we can consider the following PNU formulation \citep{pmlr-v70-sakai17a} using a weight $0\leq \omega \leq 1$ such that $\widehat{J}^{\mathrm{PNUTW}}(g) = \omega \widehat{J}^{\mathrm{PUTW}}(g) + (1-\omega) \widehat{J}^{\mathrm{TW}}(g)$. We call this approach as PNUTW. \citet{Ktena2019} also proposed PU Learning for DF learning, but their formulation is different from ours. As we explain in \ref{rem:pulearning}, their formulation provides an estimator of the biased conditional probability defined as $\frac{\gamma}{\zeta}p\left(Y_i(E^T)\mid X_i\right)$.

\citet{yasui2020} proposed an Importance Weighting (IW) based method called FSIW using the stationarity assumption and a similar variation as section \ref{sec:relation_randomvar}, they obtain 
\begin{align*}
&\frac{p(C_i=+1\mid X_i, E^t_{i})}{p(Y_i(E^t_{i})= +1 \mid X_i, E^t_{i})}= p(S_i(E^t_{i}) = +1\mid C_i = +1, X_i, E^t_{i})^{-1}
\end{align*}
and 
\begin{align*}
&\frac{p(C_i=-1\mid X_i, E^t_{i})}{p(Y_i(E^t_{i}) = -1 \mid X_i, E^t_{i})}= 1 - \frac{p(S_i(E^t_{i}) = -1, C_i = +1\mid X_i, E^t_{i})}{p(Y_i(E^t_{i}) = +1 \mid X_i, E^t_{i})}.
\end{align*}
Then, they define an empirical risk with IW as
\begin{align*}
\widehat{J}^{\mathrm{FSIW}}(g) = \frac{1}{N}\sum_{i\in\mathcal{D}} \ell\Big(Y_i(E^t_{i})g(X_i)\Big)\hat{r}\Big(Y_i(E^t_{i}), X_i, E^t_{i}\Big),
\end{align*}
where $\hat{r}(y,x, e)$ is an estimator of 
\[r(y, x, e) = \frac{p(C_i=y\mid X_i=x, E^t_{i}=e)}{p(Y_i(e) = y \mid X_i=x, E^t_{i}=e)}.\] Note that the empirical risk of the FSIW is not unbiased, but is consistent with $J(g)$. Compared with FSIW, our proposed method approximates the same risk under the same assumptions but has two preferable features: the variance of FSIW tends to be larger owing to the density ratio estimation; that is, convDF allows us to minimize the loss directly while FSIW requires a two-step procedure.

Finally, we introduce the delayed feedback model (DFM). \citet{Chapelle2014} specified the models of $p(C_i\mid X_i)$ and $p(D_i\mid X_i, C_i = +1)$ as $p(C_i\mid X_i) = \frac{1}{1+\exp(-g(X_i))}$ and $p(D_i = d\mid X_i, C_i = +1) = \lambda(X_i)\exp(-\lambda(X_i)d)$, where $D_i$ denotes the periods after observing the sample until observing the positive label and the function $\lambda(X_i)$ is called the hazard function in survival analysis. \citet{Chapelle2014} used $\lambda(x) = \exp(h(x))$ by using a function $h:\mathcal{X}\to \mathbb{R}$. Regarding the models $g(x)$ and $h(x)$, \citet{Chapelle2014} proposed linear models, and only $g(x)$ is used to predict. We summarise the comparison of these methods in this section in Table~\ref{tab:comparison}.
\begin{figure}
\centering
    \includegraphics[width=88mm]{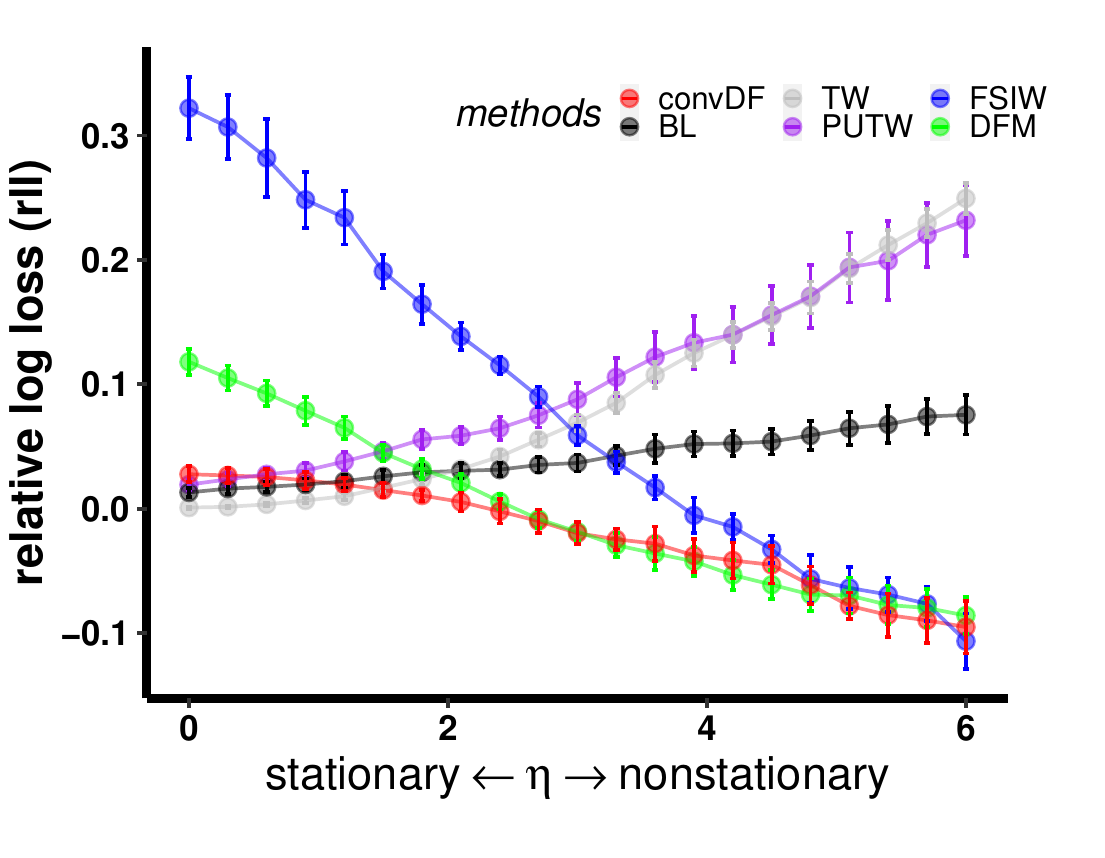}
    \caption{The average relative log loss (rll) of $20$ iterations of synthetic data experiment. The error bar shows the $95\%$ confidence interval. The result of nnDF is omitted because the results are the almost the same with convDF.}
    \label{fig:synthetic}
\end{figure}
\begin{table*}[!ht]
\caption{Negative log loss (nLL), accuracy (ACC), and area under the precision-recall curve (AUC). The best performing methods, except for OracleLogistic, are in bold.}
\label{tbl:results}
\begin{center}
\scalebox{1}{
\begin{tabular}{|l|rrr|rrr|rrr|rrr|}
\hline
{} & \multicolumn{3}{|c|}{Day 54} & \multicolumn{3}{c|}{Day 55} & \multicolumn{3}{c|}{Day 56} & \multicolumn{3}{c|}{Day 57}\\ 
\hline
{} &      nLL &      ACC &      AUC &      nLL &      ACC &      AUC &      nLL &      ACC &      AUC &      nLL &      ACC &      AUC \\
\hline
nnDF &  0.265 &  0.935 &  0.817 &  \textbf{0.269} &  \textbf{0.929} &  0.829 &  \textbf{0.283} &  \textbf{0.917} &  0.842 &  \textbf{0.326} &  0.888 &  0.815 \\
BL &  0.290 &  \textbf{0.936} &  0.864 &  0.314 &  \textbf{0.929} &  0.859 &  0.355 &  \textbf{0.917} &  0.838 &  0.440 &  0.888 &  0.797 \\
TW &  \textbf{0.260} &  \textbf{0.936} &  0.883 &  0.284 &  \textbf{0.929} &  \textbf{0.878} &  0.324 &  \textbf{0.917} &  \textbf{0.858} &  0.416 &  0.888 &  0.822 \\
PUTW &  0.320 &  0.889 &  \textbf{0.889} &  0.368 &  0.928 &  0.874 &  0.335 &  0.917 &  0.856 &  0.441 &  0.888 &  0.809 \\
FSIW &  0.274 &  \textbf{0.936} &  0.869 &  0.300 &  \textbf{0.929} &  0.862 &  0.340 &  \textbf{0.917} &  0.840 &  0.374 &  \textbf{0.908} &  \textbf{0.827} \\
DFM &  0.280 &  \textbf{0.936} &  0.867 &  0.320 &  \textbf{0.929} &  0.860 &  0.356 &  \textbf{0.917} &  0.839 &  0.444 &  0.888 &  0.794 \\
\hline
Oracle &  0.070 &  0.998 &  1.000 &  0.068 &  0.999 &  1.000 &  0.069 &  0.999 &  1.000 &  0.078 &  0.999 &  0.999 \\
\hline
\end{tabular}
}
\end{center}

\begin{center}
\scalebox{1}{
\begin{tabular}{|l|rrr|rrr|rrr||rrr|}
\hline
{} & \multicolumn{3}{|c|}{Day 58} & \multicolumn{3}{c|}{Day 59} & \multicolumn{3}{c||}{Day 60} & \multicolumn{3}{c|}{Average}\\ 
\hline
{} &      nLL &      ACC &      AUC &      nLL &      ACC &      AUC &      nLL &      ACC &      AUC &      nLL &      ACC &      AUC \\
\hline
nnDF &  0.653 &  \textbf{0.763} &  0.484 &  0.421 &  0.781 &  0.904 &  0.233 &  0.983 &  \textbf{0.994} &  0.347 &  0.888 &  0.810 \\
BL &  0.589 &  \textbf{0.763} &  0.681 &  0.340 &  0.800 &  0.975 &  0.281 &  0.826 &  0.990 &  0.371 &  0.867 &  0.859 \\
TW &  0.603 &  \textbf{0.763} &  0.668 &  0.518 &  0.763 &  0.847 &  0.472 &  0.753 &  0.923 &  0.408 &  0.852 &  0.846 \\
PUTW &  \textbf{0.570} &  0.762 &  0.617 &  0.662 &  0.763 &  0.607 &  0.735 &  0.753 &  0.611 &  0.487 &  0.845 &  0.802 \\
FSIW &  0.582 &  \textbf{0.763} &  \textbf{0.688} &  \textbf{0.208} &  \textbf{0.958} &  \textbf{0.995} &  \textbf{0.142} &  \textbf{0.995} &  0.993 &  \textbf{0.312} &  \textbf{0.916} &  \textbf{0.868} \\
DFM &  0.589 &  \textbf{0.763} &  0.684 &  0.315 &  0.821 &  0.979 &  0.256 &  0.840 &  0.992 &  0.365 &  0.872 &  0.858 \\
\hline
Oracle &  0.144 &  0.998 &  0.997 &  0.118 &  0.996 &  0.995 &  0.110 &  0.995 &  0.994 &  0.093 &  0.998 &  0.998 \\
\hline
\end{tabular}
}
\end{center}
\end{table*}

\section{Experiments}
In this section, we compared our proposed methods to other existing methods in two types of experiments. In the first experiment, we used a synthetic dataset to show that the performance of the model corresponding to the DF learning varies with the stationarity of the data. For the second experiment, we used a dataset provided by \citet{Chapelle2014} \footnote{\url{https://labs.criteo.com/2013/12/conversion-logs-dataset} }\footnote{The same dataset is available in the following GitHub repository: \url{https://github.com/ThyrixYang/es_dfm}.} to compare the methods in real-world data. 

\subsection{Synthetic Data}
In this experiment, we generate a dataset that mimics the setting in online advertising. We assume the label is a CV which is the purchase of some items after clicking the ad.
The length of the generated data is eight days, and we use the first seven days as the training data and the rest as test data. The number of samples per day is $4800$, and each sample is assigned a timestamp.
To see the impact of the stationary assumption, we add data that the stationarity assumption is not met as the days go on. This often happens in advertising platforms, such as when a new campaign is launched. Since there are fewer data of the new campaigns, the test data has some shift from the training data, so the stationarity assumption does not hold in this setting.

For the experiment, we need to generate data containing $C$, $X$, and $Y(E^T)$. We generate $C_i$ with the probability $P(C \mid X) = \frac{1}{1+\exp(-X\alpha)}$, where $\alpha$ is the parameter we randomly decide. We draw delay time from the one-sided normal distribution with standard deviation $\sigma = X\beta$, where $\beta$ is the parameter we randomly generate. If the sum of randomly assigned arriving time and delay time exceeds the duration of training data, then we set $S = 1$ for such a sample. By using randomly assigned arriving time, delay time, and $C$, we set $Y(E^T)$. For $X$, there are $20$ of binary features and $7$ of campaign binary features. The binary features are drawn from the binomial distribution. For five features, the expected value is determined by the uniform distribution whose range is $0.1$ to $0.3$, and the remaining 15 features are also randomly determined from $0.3$ to $0.7$. The true parameters for CVR and Delay of these binary features, which are $\alpha$ and $\beta$, are sampled from the uniform distribution from $-0.5$ to $0.5$ and $0$ to $10$ accordingly. For the campaign feature, we decide its value as follows.
From day $1$ to day $7$, a new campaign is added every day. The true CVR of the campaigns is set to be larger the later they are added. This is expected to degrade the performance of TW since the training data of TW may miss some data for the campaign, which is added after the time window. By contrast, the methods for DF learning can use the data of the new campaign, so we expect these methods to perform better than TW. To show the influence of the stationarity, we introduce the shift parameter $\eta$ and then define the parameter of campaigns added at day $d$ for CVR as $\alpha_{camp_{d}} = \frac{d}{7}\eta$. If we set $\eta = 0$, TW will be the best model since there is no data shift and the stationarity assumption is satisfied. Note that the campaign has no effect on the delay in this dataset.

We compare the proposed method (convDF) with BL, TW, PUTW, FSIW, and DFM. For all methods except DFM, we use logistic loss. We also train a logistic regression model with ground truth label $C_i$ and call this model as Oracle. Note that Oracle method is ideal and unrealizable since we do not have access to $C_i$. To train nnDF, we use a plain gradient descent algorithm, not the descent/ascent algorithm. For each model, we use a linear model with the $D = 28$ dimensional feature and $L_2$ regularization defined as $R(g) := \frac{1}{D}\sum^{D}_{d=1}\|\theta_d\|^2_2$, where $\theta_d$ is the $d$-th parameter of the linear model and $\|\cdot\|_2$ is the $L_2$ norm. For each method, we used Optuna\footnote{\url{https://optuna.readthedocs.io/en/stable/}} software to tune the hyper-parameter. Its search range is $10e^{-6}$ to $10e^{-1}$. Since the average values of $Y$ and $C$ change by each trial, we measure the performance by relative log loss as $rll = \frac{logloss - logloss_{Oracle}}{logloss_{Oracle}}$, where $logloss_{Oracle}$ is the logistic loss of Oracle. 

The average result of 20 trials is shown in Figure \ref{fig:synthetic}. In general, the performance of FSIW, DFM, and convDF improves, and the performance of BL, TW, and PUTW deteriorates as the $\eta$ increases. Note that if $\eta$ is high, the rll is negative since the methods of DF learning can use the most recent data to improve performance. The best-performing model changes by the value of $\eta$. For instance, TW performs the best when $\eta$ is lower. This is mainly because the stationary assumption is satisfied, so campaigns not included in the training data of TW have similar CVR to other campaigns. On the other hand, when $\eta$ is large, the performance of either convDF, FSIW, or DFM is the best. Here, convDF performs the best in the middle and well in most values of $\eta$. This result shows the practical advantage of the proposed method since it is not possible to know the strength of the stationarity before the model training in practice.

\subsection{Criteo Dataset}
In the second experiment, we used the real-world dataset provided by Criteo as \citet{Chapelle2014}. This dataset contains the click and conversion log data of multiple campaigns. We followed the experimental setting and feature engineering of \citet{Chapelle2014} to provide a fair comparison. We separate the original dataset into seven datasets as follows. There are $7$ days of test data, and for each test day, a model is trained using the previous $3$ weeks of data. Each training set contains slightly less than $6 M$ examples. All features are mapped into a $2^{24}$ sparse binary feature vector via the hashing trick \citep{weinberger2009}. As in the first experiment, we use a linear model and $L_2$ regularization, but the dimension of the feature is $2^{24}$.
Regarding metrics, we used the negative log loss (nLL), accuracy (ACC), and the area under the precision-recall curve (AUC). In online ads, the estimated probability of $p(C_i\mid X_i)$ is essential for deciding the bidding price. Therefore, the nLL is the most important metric in such an application. 

We compare nnDF with the same methods used in the first experiment. Note that we do not report the result of convDF, since it diverged. For each method, we choose a regularization parameter $\lambda$ from the set $\{0.1, 0.05, 0.01, 0.005\}$ using two-fold cross-validation. We present the experimental results in Table~\ref{tbl:results}. The results of each day and an average of $7$ days using the test data over $7$ days are presented. As in the first experiment, the assumption of stationarity is not met in the Criteo Dataset, as campaigns are added as the days go by. Especially, CVR for campaigns added after Day58 tends to be higher than those added before that date. As a result, the strength of the non-stationarity varies each day, so the best-performing methods are different for these $7$ days. While nnDF performs the best for days 55 to 57, FSIW performs the best for days 59 and 60 and also on the average of 7 days. Note that the average result depends on how much highly non-stationary data is included.

\section{Conclusion}
In this paper, we propose novel methods for DF learning under time window and stationarity assumptions. The basic formulation of the proposed methods employs a convex unbiased empirical risk estimator. We also determine the estimation error bounds of the proposed methods. Finally, we demonstrate that our proposed method performs well for any strength of stationarity, but other methods perform well for certain strengths while performing poorly for others.

A promising extension of the proposed methods involves online and continuous learning. Herein, we only discuss a general formulation for DF learning and do not develop a method involving online learning. However, because our formulation is simple, convex, and easy to optimize, we consider that it should not be difficult to develop an online learning method with theoretical guarantees.
\bibliographystyle{ACM-Reference-Format}
\bibliography{reference}

\appendix
\newpage

\twocolumn

\def\thesection{Appendix \Alph{section}}
\section{Gradients of convDF and nnDF with Logistic Loss}
\label{appdx:gradients}
Here, we show the gradients of $\widehat{J}_{\mathrm{logistic}}(g)$ and $\widetilde{J}_{\mathrm{logistic}}(g)$. For simplicity, we assume a linear model for the model $g(X_i)$; that is, for a $D$-dimensional $X_i=(X_{i,1},X_{i,2}\dots,X_{i,D})^\top$, the model is given as $g(X_i)=\sum^D_{d=1}\theta_dX_{i,d}$\footnote{Suppose that the bias term is included in $X_i$.}, where $\theta = (\theta_1, \theta_2, \dots, \theta_D)^\top$ is a parameter of the linear model and $\theta_d\in\mathbb{R}$. Let us redefine the convDF and nnDF risks as $\widehat{J}_{\mathrm{logistic}}(g, \theta)$ and $\widetilde{J}_{\mathrm{logistic}}(g, \theta)$, respectively. Then, the gradients of $\widehat{J}_{\mathrm{logistic}}(g, \theta)$ is given as follows:
\begin{align*}
\frac{\partial \widehat{J}_{\mathrm{logistic}}(g, \theta)}{\partial \theta} =& \frac{1}{N}\sum_{i\in\mathcal{D}}\big(\mathbbm{1}[Y_i = +1] - \psi(X_i)\big)X_i \\ 
& - \frac{1}{M}\sum_{j\in\mathcal{E}}\mathbbm{1}\left[\Big(S_j(E^{T-\tau}_{j} )=-1\Big) \land \Big(C_j=+1\Big)\right] X_{j},
\end{align*}
where $\psi(X_i) = \frac{1}{1+\exp\left(g(X_i)\right)}$. The gradients of $\widetilde{J}_{\mathrm{logistic}}(g, \theta)$ with a plain gradient/descent method is $ \frac{\partial \widehat{J}_{\mathrm{logistic}}(g, \theta)}{\partial \theta}$ when $\widehat{J}^{(-)}_{\mathcal{D}}(g) - \widehat{J}^{(-)}_{\mathcal{E}}(g) > 0$. Otherwise, the gradient of $\widetilde{J}_{\mathrm{logistic}}(g, \theta)$ is:
\begin{align*}
    \frac{\partial \widetilde{J}_{\mathrm{logistic}}(g, \theta)}{\partial \theta} &= \frac{\mathbbm{1}[Y_i = +1]}{N} \sum_{i\in\mathcal{D}} (1 - \psi(X_i))X_i \\ 
    +& \frac{1}{M}\mathbbm{1}\left[(S_j(E^{T-\tau}_{j} )=-1) \land (C_j=+1)\right] \sum_{j\in\mathcal{E}}(1 - \psi(X_j))X_j.
\end{align*}

\section{Biased PU Learning}
\label{rem:pulearning}
This problem arises in various practical situations, such as information retrieval and outlier detection \citep{elkan2008learning, ward2009presence, pmlr-v5-scott09a, blanchard2010semi, li2009positive,nguyen2011positive}. In PU learning, there are censoring and case-control scenarios \citep{elkan2008learning}. The \emph{convex PU learning} \citet{IEICE:duPlessis+Sugiyama:2014,duplessis2015} is a method for case-control scenario, which constructs \emph{unbiased and convex estimator} of the true classification risk. By using the methods proposed by \citet{duplessis2015} and \citet{kiryo2017}, \citet{Ktena2019} proposed minimizing $\widehat{J}^{\mathrm{PU}}(g) = \frac{1}{N}\sum_{i\in\mathcal{D}} \ell\big(-g(X_i)\big)+\hat{\gamma}\frac{\sum_{i\in\mathcal{E}} \mathbbm{1}\left[Y_i\left(E^t_{i}\right)=+1\right]\left(\ell\big(g(X_j)\big) - \ell\big(-g(X_j)\big)\right)}{\sum^N_{i=1} \mathbbm{1}\left[Y_i\left(E^t_{i}\right)=+1\right]}$, where $\hat{\gamma}$ is a parameter estimated by the \emph{class-prior estimation} \citep{christoffel2016class,pmlr-v48-ramaswamy16,jain2016nonparametric,1809.05710}. However, as \citet{1809.05710} and \citet{kato2018learning} showed, the minimizer of the population version of the empirical risk matches the biased probability $\frac{\gamma}{\zeta}p\left(Y_i(E^T)\mid X_i\right)$, i.e., the empirical minimization is the same as the naive logistic regression.

\section{Proofs of Theoretical Analysis}

Firstly, we introduce the McDiarmid’s inequality \citep{mcdiarmid1989method}.
\begin{proposition}[McDiarmid’s Inequality \citep{mcdiarmid1989method,Sammut2010}]
Suppose $f:\mathcal{X}^n\to\mathbb{R}$ satisfies the bounded differences property. That is, for all $i=1,\dots,n$, there is a $c_i\geq 0$ such that, for all $x_1,\dots, x_n,x'\in\mathcal{X}$, $\big| f(x_1,\dots, x_n) - f(x_1,\dots, x_{i-1}, x', x_{i+1}, \dots, x_n)\big|\leq c_i$.
If $X=(X_1,\dots, X_n)\in\mathcal{X}^n$ is a random variable drawn according to $P^n$ and $\mu=\mathbb{E}_{P^n}[f(X)]$, then, for all $\epsilon > 0$, $P^n\big(f(X)-\mu \geq \epsilon \big)\leq \exp\left(\frac{2\epsilon^2}{\sum^n_{i=1}c^2_i}\right)$.
\end{proposition}

For dealing with the non-negative correction, we define the following consistent correction function, which includes the non-negative correction as a special case. Then, following \citet{kiryo2017} and \citet{kato2020dre}, we define the following alternative version of Rademacher complexity \citep{Bartlett2003} for bounding the estimation error.

\begin{definition}[Consistent correction function \cite{LuMitigating2020}]
A function \(\rho : \Re \to \Re\) is called a consistent correction function if it is Lipschitz continuous, non-negative and \(\rho(x) = x\) for all \(x \geq 0\).
\label{def:consistent-correction}
\end{definition}
\begin{definition}[Rademacher complexity]
Given \(n \in \mathbb{N}\) and a distribution \(p\), define the Rademacher complexity \(\Radnp(\mathcal{F})\) of a function class \(\mathcal{F}\) as $\Radnp(\rClass) := \E_p\ERad\left[\sup_{f\in\mathcal{F}}\left|\frac{1}{n} \sum_{i=1}^n \rad_i f(\X_i)\right|\right]$, where \(\{\sigma_i\}_{i=1}^n\) are Rademacher variables (i.e., independent variables following the uniform distribution over \(\{-1, +1\}\)) and \({\{\X_i\}_{i=1}^n \overset{\text{i.i.d.}}{\sim} p}\).
\label{def:rademacher-complexity}
\end{definition}

Secondly, we introduce a useful proposition on symmetrization with consistent correction function from \citet{kato2020dre}.

\begin{proposition}[Symmetrization under Lipschitz-continuous modification, \citep{kato2020dre}]
\label{lmm:sym_lip}
Let \(0 \leq a < b\), \(J \in \Na\), and \(\{K_j\}_{j=1}^J \subset \Na\).
Given i.i.d. samples \(\Xsetjk := \{X_i\}_{i=1}^{\njk}\) each from a distribution \(\pjk\) over \(\InSpace\),
consider a stochastic process \(\hS\) indexed by \(\F \subset (a, b)^\mathcal{X}\) of the form $\hS(\f) = \sum_{j=1}^J \rho_j\left(\sumjk\hEjk [\ljk(\f(\X))]\right)$, where each \(\rho_j\) is a \(\Liprhoj\)-Lipschitz function on \(\Re\),
\(\ljk\) is a \(\LipLossjk\)-Lipschitz function on \((a, b)\),
and \(\hEjk\) denotes the expectation with respect to the empirical measure of \(\Xsetjk\).
Denote \(\EhS(\f) := \EX \hS(\f)\) where \(\EX\) is the expectation with respect to the product measure of \(\{\Xsetjk\}_{(j, k)}\).
Here, the index \(j\) denotes the grouping of terms due to \(\rho_j\), and \(k\) denotes each sample average term.
Then we have $\EX \supf |\hS(\f) - \EhS(\f)| \leq 4 \sumj \sumjk \Liprhoj \LipLossjk \mathcal{R}^{p_{(j,k)}}_{n_{(j,k)}}(\F)$.
\label{lem:general-symmetrization}
\end{proposition}

\subsection{Proof of Lemma~\ref{lmm:bias_consistency}}
The procedure of the proof mainly follows \citet{kiryo2017}. 
Let $F(\mathcal{D}, \mathcal{E})$ be the cumulative distribution function of $(\mathcal{D}, \mathcal{E})$. Given the above definitions, the measure of $\mathcal{B}(g)$ is defined by $\mathrm{Pr}\left(\mathcal{B}(g)\right) = \int_{(\mathcal{D}, \mathcal{E})\in\mathcal{B}(g)} dF(\mathcal{D}, \mathcal{E})$. Since $\widetilde{J}(g)$ is identical to $\widehat{J}(g)$ on $\mathcal{A}(g)$ and different from $\widehat{J}(g)$ on $\mathcal{B}(g)$, we have $\mathrm{Pr}\left(\mathcal{B}(g)\right) = \mathrm{Pr}\left(\widetilde{J}(g)\neq \widehat{J}(g)\right)$. This result means that the measure of $\mathcal{B}(g)$ is non-zero if and only if $\widetilde{J}(g)$ differs from $\widehat{J}(g)$ with a non-zero probability.

Based on the facts that $\widehat{J}(g)$ is unbiased and $\widetilde{J}(g) - \widehat{J}(g) = 0$ on $\mathcal{A}(g)$, we have $\mathbb{E}\left[\widetilde{J}(g)\right] - J(g) = \mathbb{E}\left[\widetilde{J}(g) - \widehat{J}(g)\right]= \int_{(\mathcal{D}, \mathcal{E})\in \mathcal{A}(g)} \widetilde{J}(g) - \widehat{J}(g) \mathrm{d}F(\mathcal{D}, \mathcal{E}) + \int_{(\mathcal{D}, \mathcal{E})\in \mathcal{B}(g)} \widetilde{J}(g) - \widehat{J}(g) \mathrm{d}F(\mathcal{D}, \mathcal{E}) = \int_{(\mathcal{D}, \mathcal{E})\in \mathcal{B}(g)} \widetilde{J}(g) - \widehat{J}(g) dF(\mathcal{D}, \mathcal{E})$. As a result, $\mathbb{E}\left[\widetilde{J}(g)\right] - J(g) > 0$ if and only if $\int_{(\mathcal{D}, \mathcal{E})\in\mathcal{B}^{(-)}(g)} dF(\mathcal{D}, \mathcal{E}) > 0$ due to the fact $\widetilde{J}(g) - \widehat{J}(g > 0$ on $\mathcal{B}(g)$. That is, the bias of $\widetilde{J}(g)$ is positive if and only if the measure of $\mathcal{B}(g)$ is non-zero.

We prove Eq.~\eqref{eq:bias_prob_bound} by the method of bounded difference, for that $\mathbb{E}\left[\widehat{J}^{(-)}_{\mathcal{D}}(g) - \widehat{J}^{(-)}_{\mathcal{E}}(g) \right] = J^{(-)}(g)\geq \alpha$. We have assumed that $0\leq \ell(\cdot)\leq C_\ell$, and thus the change of $\widehat{J}^{(-)}(g)$ will be no more than $C_\ell/N$ if some $X_i\in \mathcal{X}_{\mathcal{D}/\mathcal{E}}$ is replaced, or the change of $\widehat{J}^{(-)}(g)$ will be no more than $C_\ell/N + C_\ell/M$ if some $X_i\in \mathcal{X}_{\mathcal{E}}$ is replaced. Subsequently, McDiarmid's inequality \citep{mcdiarmid1989method} implies 
\begin{align*}
    &\mathrm{Pr}\left(J^{(-)}(g) - \left(\widehat{J}^{(-)}_{\mathcal{D}}(g) - \widehat{J}^{(-)}_{\mathcal{E}}(g)\right)\geq \alpha\right)\\
    &\leq \exp\left(-\frac{2\alpha^2}{\big(N-M\big)\left(C_\ell/N\right)^2 + M\left(C_\ell/N + C_\ell/M\right)^2}\right)\\
    &= \exp\left(-\frac{2\alpha^2 / C^2_\ell}{3/N + 1/M}\right).
\end{align*}
Taking into account that 
\begin{align*}
\mathrm{Pr}\left(\mathcal{B}(g)\right) &= \mathrm{Pr}\left(\widehat{J}^{(-)}_{\mathcal{D}}(g) - \widehat{J}^{(-)}_{\mathcal{E}}(g) < 0\right)\\
&\leq \mathrm{Pr}\left(\widehat{J}^{(-)}_{\mathcal{D}}(g) - \widehat{J}^{(-)}_{\mathcal{E}}(g) \leq J^{(-)}(g) - \alpha\right)\\
&= \mathrm{Pr}\left(J^{(-)}(g) - \left(\widehat{J}^{(-)}_{\mathcal{D}}(g) - \widehat{J}^{(-)}_{\mathcal{E}}(g)\right)\geq \alpha\right),
\end{align*}
we complete the proof.

\subsection{Proof of Theorem~\ref{thm:bias_consistency}}
It has been proven in Lemma~\ref{lmm:bias_consistency} that $\mathbb{E}\left[\widetilde{J}(g)\right] - J(g)= \int_{\mathcal{X}\in \mathcal{B}^{(-)}(g)} \widetilde{J}(g) - \widehat{J}(g) dF(\mathcal{X})$, thus the exponential decay of the bias is obtained via 
\begin{align*}
    &\mathbb{E}\left[\widetilde{J}(g)\right] - J(g)\\ 
    &\leq \sup_{(\mathcal{D}, \mathcal{E})\in \mathcal{B}^{(-)}(g)}\left(\widetilde{J}(g) - J(g)\right)\cdot \int_{\mathcal{X}\in \mathcal{B}^{(-)}(g)}dF(\mathcal{X})\\
    &\leq \sup_{(\mathcal{D}, \mathcal{E})\in \mathcal{B}^{(-)}(g)}\left(\widehat{J}^{(-)}_{\mathcal{E}}(g) - \widehat{J}^{(-)}_{\mathcal{D}}(g)\right)\cdot \mathrm{Pr}\left(\mathcal{B}^{(-)}(g)\right)\leq C_\ell \Delta_g.
\end{align*}
The deviation bound Eq.~\eqref{eq:dev_bound} is due to 
\begin{align*}
    \left|\widetilde{J}(g) - J(g)\right| &\leq \left|\widetilde{J}(g) - \mathbb{E}\left[\widetilde{J}(g)\right]\right| + \left|\mathbb{E}\left[\widetilde{J}(g)\right] - J(g)\right| \\
    &\leq \left|\widetilde{J}(g) - \mathbb{E}\left[\widetilde{J}(g)\right]\right| + C_\ell \Delta_g .
\end{align*}

The change of $\widetilde{J}(g)$ will be no more than $2C_\ell/N$ if some $X_i\in\mathcal{X}_{\mathcal{D}/\mathcal{E}}$ is replaced, or it will be no more than $2C_\ell/N + 2C_\ell/M$ if some $X_i\in\mathcal{X}_{\mathcal{D}/\mathcal{E}}$ is replaced. Therefore, McDiarmid's inequality gives us

\begin{align*}
    &\mathrm{Pr}\left\{ \left|\widetilde{J}(g) - \mathbb{E}\left[\widetilde{J}(g)\right]\right|\geq \epsilon\right\} \\
    &\leq 2\exp\left(-\frac{2\epsilon^2}{\big(N-M\big)\left(2C_\ell/N\right)^2 + M\left(2C_\ell/N + 2C_\ell/M\right)^2}\right) \\
    &= 2\exp\left(-\frac{\epsilon^2 / C^2_\ell}{6/N + 2/M}\right)
\end{align*}

or equivalently, with probability at least $1-\delta$,
\begin{align*}
    &\left|\widetilde{J}(g) - \mathbb{E}\left[\widetilde{J}(g)\right]\right| \le_\ell\sqrt{\big(6/N + 2/M\big)\log \big(2/\delta\big)}\\
    &= C_\ell\sqrt{2\log \big(2/\delta\big)}\sqrt{\big(3/N + 1/M\big)}\leq C_\ell\sqrt{2\log \big(2/\delta\big)}\left(\sqrt{3/N} + \sqrt{1/M}\right).
\end{align*}
On the other hand, the deviation bound Eq.~\eqref{eq:dev2} is obtained from $\left|\widetilde{J}(g)- J(g)\right|\leq \left|\widetilde{J}(g)- \widehat{J}(g)\right| + \left|\widehat{J}(g)- J(g)\right|$,
where $\left|\widetilde{J}(g)- \widehat{J}(g)\right|$ with probability at most $\Delta_g$, and $\left|\widehat{J}(g)- J(g)\right|$ has the same bound with $\left|\widetilde{J}(g) - \mathbb{E}\left[\widetilde{J}(g)\right]\right|$.

\subsection{Proof of Theorem~\ref{thm:est_error_convDF}}

Since $\hat{g}$ minimizes $\widehat{J}^\omega(g)$, we have 
\begin{align*}
    J(\hat{g}) - J(g^*) &= J(\hat{g}) - \widehat{J}(\hat{g}) + \widehat{J}(\hat{g}) - J(g^*) \leq J(\hat{g}) - \widehat{J}(\hat{g}) + \widehat{J}(g^*) - J(g^*)\\ 
    &\leq 2 \sup_{g\in\mathcal{H}} \left| \widehat{J}(g) - J(g) \right|.
\end{align*}
By applying McDiarmid's inequality to $\sup_{g\in\mathcal{H}} \left| \widehat{J}(g) - J(g) \right|$, 
\begin{align*}
\mathrm{Pr}\left(\sup_{g\in\mathcal{H}} \left| \widehat{J}(g) - J(g) \right| - \mathbb{E}\left[\sup_{g\in\mathcal{H}} \left| \widehat{J}(g) - J(g) \right|\right]\geq \epsilon \right) &\leq  \exp\left(-\frac{\varepsilon^2 / C^2_\ell}{6/N + 2/M}\right)
\end{align*}
or equivalently, the following bound holds with probability at least $1-\delta$:
\begin{align*}
&\sup_{g\in\mathcal{H}} \left| \widehat{J}(g) - J(g) \right|\leq \annot{\mathbb{E}\left[\sup_{g\in\mathcal{H}} \left| \widehat{J}(g) - J(g) \right|\right]}{Expected maximal deviation} + C_\ell\sqrt{\big(6/N + 2/M\big)\log \big(1/\delta\big)}.
\end{align*}
By using Proposition~\ref{lmm:sym_lip} for $\rho(x) = x$, 
\begin{align*}
    \mathbb{E}\left[\sup_{g\in\mathcal{H}} \left| \widehat{J}(g) - J(g) \right|\right]\leq 4C_\ell \mathcal{R}_\mathcal{D}(\mathcal{H}) + 4C_\ell \mathcal{R}_\mathcal{E}(\mathcal{H}).
\end{align*}

\subsection{Proof of Theorem~\ref{thm:est_error_nnDF}}
\begin{proof}
Since $\hat{g}$ minimizes $\widehat{J}^\omega(g)$, we have
\begin{align*}
 J(\hat{g}) - J(g^*) &= J(\hat{g}) - \widetilde{J}(\hat{g}) + \widetilde{J}(\hat{g}) - J(g^*) \\
 &\leq J(\hat{g}) - \widetilde{J}(\hat{g}) + \widetilde{J}(g^*) - J(g^*) \leq 2 \sup_{g\in\mathcal{H}} \left| \widetilde{J}(g) - J(g) \right| \\
 &= \annot{2 \sup_{g\in\mathcal{H}} \left| \widetilde{J}(g) - \mathbb{E}\left[\widetilde{J}(g)\right] \right|}{Maximal deviation} + \annot{2 \sup_{g\in\mathcal{H}} \left| \mathbb{E}\left[\widetilde{J}(g)\right] - J(g) \right|}{Bias}.
\end{align*}

For applying McDiarmid's inequality to $\sup_{g\in\mathcal{H}} \left| \widehat{J}(g) - J(g) \right|$,
\begin{align*}
\mathrm{Pr}\left(\sup_{g\in\mathcal{H}} \left| \widetilde{J}(g) - \mathbb{E}\left[\widetilde{J}(g)\right]  \right| - \mathbb{E}\left[\sup_{g\in\mathcal{H}} \left| \widehat{J}(g) - J(g) \right|\right]\geq \epsilon \right) = \exp\left(-\frac{\varepsilon^2 / C^2_\ell}{6/N + 2/M}\right)
\end{align*} 
or equivalently, the following bound holds with probability at least $1-\delta$:
\begin{align*}
&\sup_{g\in\mathcal{H}} \left| \widetilde{J}(g) - \mathbb{E}\left[\widetilde{J}(g)\right]  \right|\\
&\leq \annot{\mathbb{E}\left[\sup_{g\in\mathcal{H}} \left| \widetilde{J}(g) - \mathbb{E}\left[\widetilde{J}(g)\right]  \right|\right]}{Expected maximal deviation} + C_\ell\sqrt{\big(6/N + 2/M\big)\log \big(1/\delta\big)}.
\end{align*}

By using Proposition~\ref{lmm:sym_lip} for $\rho(x) = x$, we have $\mathbb{E}\left[\sup_{g\in\mathcal{H}} \left| \widehat{J}(g) - J(g) \right|\right]\leq 8C_\ell \mathcal{R}_\mathcal{D}(\mathcal{H}) + 8C_\ell \mathcal{R}_\mathcal{E}(\mathcal{H})$.

The bias term can be bounded as $\left|\widetilde{J}(g) - J(g)\right|\leq C_\ell\Delta$.
\end{proof}

\end{document}